\pdfoutput=1

\documentclass[11pt]{article}

\usepackage{acl}

\usepackage{diagbox}
\usepackage{times}
\usepackage{latexsym}
\usepackage{amsmath}
\usepackage{amssymb}
\usepackage{makecell}

\usepackage{subcaption}

\usepackage{booktabs}
\usepackage{algpseudocode}
\usepackage{algorithm}
\usepackage{microtype}

\def\arrvline{\hfil\kern\arraycolsep\vline\kern-\arraycolsep\hfilneg}
\usepackage{multirow}
\usepackage[T1]{fontenc}

\usepackage[utf8]{inputenc}

\usepackage{microtype}

\usepackage{graphicx}
\usepackage{subcaption}

\title{PoeLM: A Meter- and Rhyme-Controllable Language Model for Unsupervised Poetry Generation}

\author{Aitor Ormazabal$^{1}$ \quad Mikel Artetxe$^{2}$ \quad  Manex Agirrezabal$^{3}$ \quad Aitor Soroa$^{1}$ \quad Eneko Agirre$^{1}$  \\
$^1$HiTZ Center, University of the Basque Country (UPV/EHU) \\
$^2$Meta AI \quad $^3$University of Copenhagen \\
\texttt{\{aitor.ormazabal,a.soroa,e.agirre\}@ehu.eus} \\ 
\texttt{artetxe@meta.com} \quad \texttt{manex.aguirrezabal@hum.ku.dk} \\
}

\begin{document}
\maketitle
\begin{abstract}

Formal verse poetry imposes strict constraints on the meter and rhyme scheme of poems. Most prior work on generating this type of poetry uses existing poems for supervision, which are difficult to obtain for most languages and poetic forms. In this work, we propose an unsupervised approach to generate poems that follow any given meter and rhyme scheme, without requiring any poetic text for training. Our method works by splitting a regular, non-poetic corpus into phrases, prepending control codes that describe the length and end rhyme of each phrase, and training a transformer language model in the augmented corpus. The transformer learns to link the structure descriptor with the control codes to the number of lines, their length and their end rhyme. During inference, we build control codes for the desired meter and rhyme scheme, and condition our language model on them to generate formal verse poetry. Experiments in Spanish and Basque show that our approach is able to generate valid poems, which are often comparable in quality to those written by humans.

\end{abstract}

\section{Introduction}

\begin{figure}[t]
\includegraphics[clip,width=\linewidth]{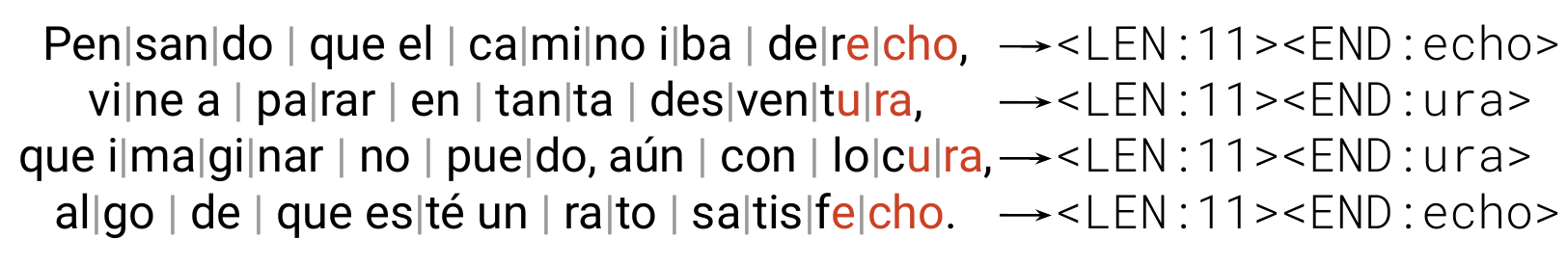}
\caption{
\textbf{A formal verse poem and its associated structure descriptor.}
The poem is the first stanza of a Spanish sonnet, which must have 4 lines of 11 syllables and follow an ABBA rhyme scheme. We use control codes to describe such constraints, and train a language model that can generate text conditioned on them.
}
\label{fig:sonnet}
\end{figure}

Despite the impressive generative capabilities of large Language Models (LMs) \citep{brown-gpt3,chowdhery2022palm,zhang2022opt} %
automatic poetry generation remains a challenging problem. \textbf{Formal verse poetry}, in particular, imposes strict constraints on the meter and rhyme scheme of poems (Figure \ref{fig:sonnet}), which cannot be directly controlled in conventional LMs.

Prior work on generating formal verse poetry has primarily focused on supervised approaches, leveraging existing poems to train LMs. This is often combined with additional techniques to impose the meter and rhyme constraints at inference time, such as using finite-state automata to discard invalid candidates \citep{ghazvininejad-etal-2016-generating}, or generating text right-to-left to better control the rhyming word \citep{lau-etal-2018-deep,jhamtani-etal-2019-learning,xue-etal-2021-deeprapper}. However, these approaches require poetic text for training, which is difficult to obtain for most languages and poetic forms.

\begin{figure*}[t]

\begin{subfigure}{\textwidth}
  \includegraphics[clip,width=\textwidth]{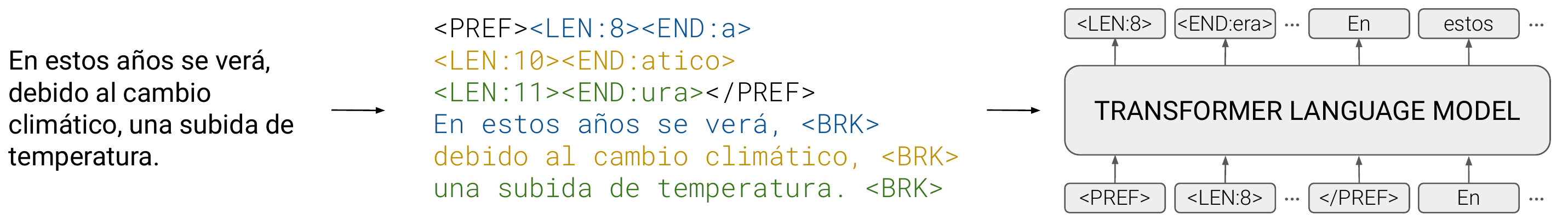}
  \caption{Training on regular, non-poetic text. Example in Spanish.}
  \label{fig:training}
\end{subfigure}
\par\bigskip
\begin{subfigure}{\textwidth}
  \includegraphics[clip,width=\textwidth]{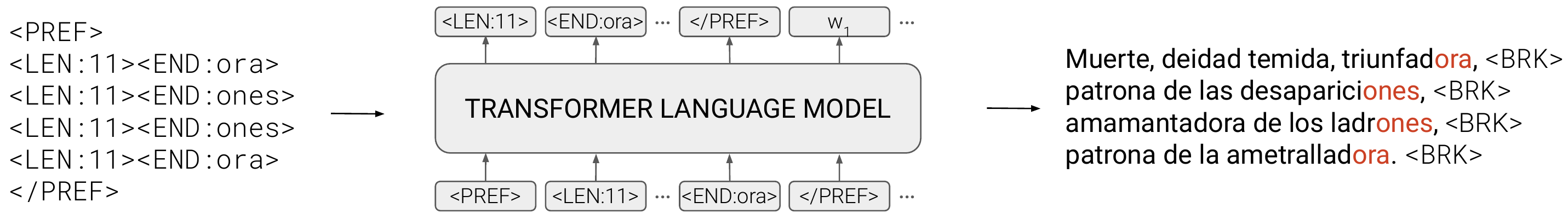}
  \caption{Generation of formal verse poetry. Example in Spanish.}
  \label{fig:inference}
\end{subfigure}

\caption{
\textbf{Proposed method.}
(a) During training, we split non-poetic text into phrases according to punctuation marks, prepend control codes describing the length and end rhyme of each phrase, and train a transformer language model on it. (b) During inference, we build a structure descriptor with control codes for the desired meter and rhyme scheme, and condition our language model on them to generate formal verse poetry.
}
\label{fig:method}
\end{figure*}

In this paper, we propose an unsupervised approach to generate formal verse poetry. Our \textbf{Poe}tic \textbf{L}anguage \textbf{M}odel (PoeLM) can be conditioned to follow any desired meter and rhyme scheme, without requiring any poem for training. As illustrated in Figure \ref{fig:method}, the key idea behind our method is that any text can be divided into phrases, which will each have a certain number of syllables and end in a certain sound that can make it rhyme with other phrases. While this structure will not follow a regular pattern for standard text, as it would for poetry, we can still annotate it automatically, and train a language model that can be conditioned on such structure descriptors. At inference time, we build a structure descriptor for the desired meter and rhyme scheme, and condition our language model on it to generate formal verse poetry. To improve results, we generate multiple candidates, which are automatically filtered and re-ranked.

Our experiments in Spanish and Basque show that our method is able to generate high quality poems meeting the desired meter and rhyme constraints, with human evaluators ranking our system higher than other humans in more than one third of the cases. Our code is available at GitHub.\footnote{\url{https://github.com/aitorormazabal/poetry_generation}}

\section{Background: formal verse poetry}
\label{sec:background}

Poetic traditions differ across languages and cultures. In this work, we focus on \textbf{formal verse poetry} in Spanish and Basque,\footnote{The selected languages where narrowed down according to the availability of publicly available high-quality syllabization and rhyme detection systems (which discarded English), as well as the fluency of the authors. } which impose strict meter and rhyme constraints as follows:
\begin{itemize}
    \item The \textbf{syllabic meter} specifies the number of lines in the poem, as well as the number of syllables %
    that each line must contain.\footnote{Some traditions impose a stress pattern in addition to the number of syllables, which is known as \textit{accentual-syllabic meter}. We do not consider this type of meter in our work, as it is not common in Spanish and Basque. %
    }
    Spanish syllabic meter allows for synalephas, where two syllables can be merged into one when one word ends in a vowel and the next starts with one. 
    For simplicity, we do not consider synalephas when counting syllables,
    although our method could easily be extended to account for them.
    \item The \textbf{rhyme scheme} specifies the pattern according to which lines must rhyme. For instance, the ABAB scheme requires the 1st line to rhyme with the 3rd one, and the 2nd line to rhyme with the 4th one. Two lines are considered to rhyme if they repeat the same sound at their last syllables.\footnote{In Spanish, two words rhyme if their sounds are identical from the last stressed vowel onwards. In Basque, two words rhyme if their sounds match from the first vowel of the second to last syllable onwards, and the following consonant groups are considered to sound the same for the purposes of rhyme: \{p,t,k\}, \{n,m\}, \{s,z,x\}, and \{b,d,g,r\}.  } %
    In addition, rhyming lines cannot end in the same word.
\end{itemize}

There are different \textbf{poetic forms} depending on the specific meter and rhyme scheme that they impose. For instance, the first stanza of a Spanish sonnet must consist of four verses with 11 syllables each, following an ABBA rhyme scheme. As illustrated in Figure \ref{fig:sonnet}, we use control codes to define such meter and rhyme constraints, which we refer to as \textbf{structure descriptors}.

\section{Proposed method}

As described in \S\ref{sec:background}, we want our system to be able to generate text that adheres to a specific structure. The key idea behind our approach is that, similar to formal verse poetry, any text adheres to a certain implicit structure. In the case of non-poetic text the structure will not follow any regular pattern, but we can still extract it and build a structure descriptor for it. We can then augment the non-poetic corpus with these descriptors, and train a regular LM on it (Figure \ref{fig:training}). The model thus learns to respect the structure provided in the descriptor, which allows us to generate formal verse poetry at inference time, by conditioning the model on the appropriate structure descriptor (Figure \ref{fig:inference}). 

We next describe the two main components of our method: structure-aware training (\S\ref{sec:pre-training}) and inference with filtered re-ranking (\S\ref{sec:re-ranking}). 

\subsection{Structure-aware training}
\label{sec:pre-training}

Let $X$ be the space of possible text sequences, and $S$ be the space of possible structure descriptors. We can define a function $s: X \rightarrow \mathcal{P}({S})$ that maps each sequence of text into its corresponding set of descriptors.\footnote{Each sequence is mapped to a subset of $S$, as the same sequence could be described by multiple descriptors.} We want to build a model that can sample from $P(X | c \in s(X))$, that is, that can sample text conditioned on an structure descriptor $c$. In theory, one could do this through rejection sampling, by repeatedly drawing sentences from $x\sim P(X)$ until one of them satisfies $c \in s(x)$. However, this is intractable in practice, since the probability of a randomly sampled text following the desired structure is practically zero. 

Instead, we train a LM that can be conditioned on any given structure (see Figure \ref{fig:training}). To that end, we start by \textbf{annotating the implicit structure} of a regular, non-poetic corpus. We first split the corpus in phrases, where we define a phrase as a sequence of text delimited by either a newline or a punctuation character (e.g., commas, colons or quotes). We do this so that the rhyme words at the end of these units correspond to natural stopping points. We then group the text in blocks of $n$ phrases, where $n$ is randomly sampled.
For each block $x$, we choose a structure descriptor $c_x\in s(x)$ that defines the length and end rhyme of each of the phrases it contains.
We then \textbf{create an augmented corpus} $(c_{x_1}, x_1, c_{x_2}, x_2, ...)$ by interleaving the previously generated structure descriptor $c_{x_i}$ before its corresponding text block $x_i$ (see Appendix \ref{sec:structure-descriptors} for more details).
Finally, we \textbf{train a transformer LM} on the augmented corpus. The control codes in the structure descriptors are treated as regular tokens, and the model is trained with the standard next token prediction objective.

\subsection{Generation with filtered re-ranking}
\label{sec:re-ranking}

At inference time, we use the LM from \S\ref{sec:pre-training} to generate formal verse poetry in 3 steps:

\paragraph{1. Candidate generation.} We specify the desired meter and rhyme scheme as a structure descriptor,\footnote{A rhyme scheme specifies which lines must rhyme, but not what the rhyme sound should be. We thus generate a concrete structure descriptor from the given scheme by sampling each rhyme sound independently from the five most common rhyme sounds in the training corpus.} and use our LM to generate text conditioned on it  (see Figure \ref{fig:inference}).
We repeat the process $k=3000$ times to generate $k$ different candidates.
In our experiments, we provide the first line of the poem to generate in addition to its structure descriptor, which is useful to define the subject and make different systems easier to compare.

\paragraph{2. Filtering.}
In practice, some of the generated candidates do not meet the given constraints or are otherwise pathological. For that reason, we filter candidates according to the following conditions:
\begin{enumerate}
    \item \textbf{\#Line.} The candidate must have the number of lines specified in the structure descriptor. %
    \item \textbf{\#Slb.} Each line must have the number of syllables specified in the structure descriptor. %
    \item \textbf{Rhyme.} Each line must end in the rhyme sound specified in the structure descriptor. %
    \item \textbf{Rep. word.} No two rhyming lines can end in the same word. %
    \item \textbf{BLEU.} In order to prevent the model from generating repetitive text, the maximum and average BLEU across any two lines must be be less than or equal to 35 and 20. %
\end{enumerate}

\paragraph{3. Re-ranking.} We score the remaining candidates for fluency using our LM, and output the one with the highest score. Different from the first step, we do not condition on the structure descriptor when doing so, which gives a measure of the general fluency. %

We test the efficacy of the second and third steps in the experiments.

\section{Experimental design}

We run experiments on Spanish and Basque. We next detail the training details (\S\ref{subsec:training}) and the automatic and human evaluation setup (\S\ref{sec:quant} and \S\ref{sec:human-evaluation}).

\subsection{Training details} \label{subsec:training}

\paragraph{Hyperparameters.} We train transformer LMs using the same settings as \citet{brown-gpt3}.  For Basque, we train a 350M model with a learning rate of $3\times 10^{-4}$ and linear decay over 300B tokens.\footnote{In practice, we stop training after seeing around 85B tokens, when performance plateaus in the validation set.}
For Spanish, we train a 760M model over 100B tokens %
using a constant\footnote{We initially planned to manually decay the learning rate according to validation perplexity. However, we did not observe performance plateauing (presumably due to the large corpus and our constrained compute budget), so the full training was done with a constant learning rate.} learning rate of $2.5 \times 10^{-4}$.

\paragraph{Corpora.} We use EusCrawl \citep{artetxe2022does} as our training corpus for Basque, which takes 2.5GB in plain text format, and a subset of 700GB from mC4 \citep{2019t5} for Spanish. Given the small size of the Basque corpus, we combine 10 versions of the corpus using different random seeds to generate the structure descriptors.

\paragraph{Preprocessing.} We use SentencePiece tokenization \citep{kudo-richardson-2018-sentencepiece} with a 50k vocabulary for each language, and reserve 8.5k tokens for the control codes in the structure descriptors. For syllabification and rhyme sound extraction we use the rules provided by \citet{agirrezabal2012finite},\footnote{ \url{https://bitbucket.org/manexagirrezabal/syllabification_gold_standard}} which are encoded as finite-state transducers implemented in Foma \cite{hulden2009foma}.

\paragraph{Models.} In addition to our proposed model (PoeLM), we train a regular LM for each language as a baseline, using the exact same hyper-parameters, tokenization, and corpora (without the interleaved structure descriptors).

\subsection{Automatic evaluation}
\label{sec:quant}

We use Spanish poems from the 20th century subset of the DISCO dataset \citep{Barbado_2021}, and Basque poems from the BDB dataset\footnote{\url{https://bdb.bertsozale.eus/}. We use the 2005 segment of the corpus.} to evaluate our approach. The DISCO and BDB datasets consist of 20k and 44k tokens before our SentencePiece tokenizer is applied, respectively. We  use the following automatic metrics:

\paragraph{Filtering rate.}
We take 10 poems\footnote{For Spanish, we use the first Stanza of full sonnets from DISCO, which consist of either 11 or 14 syllable lines, following a rhyme scheme of ABAB or ABBA. For Basque, we use \textit{Zortziko Handia} poems from BDB, which consist of 8 lines, where the odd ones are 10 syllables long, the even ones are 8 syllables long, and only the even lines are required to rhyme.} from each test set, extract the first line from them, %
generate poems for each as described in \S\ref{sec:re-ranking} following the meter and rhyme scheme of the original poem,  with $k=3000$ candidates for each,
and measure the percentage of candidates that are filtered according to the criteria in \S\ref{sec:re-ranking}. %
We compare the resulting filtering rate of our proposed PoeLM, which is conditioned on the relevant structure descriptor, %
 and a regular LM, which is not conditioned on any structure but could still generate a valid poem given enough trials.

Since, unlike PoeLM, the baseline LM does not generate break tokens to separate lines, we split the generated text into lines according to the relevant number of syllables. When this cannot be done while respecting word boundaries, we consider that the candidate is rejected for breaking the \textit{\#slb} condition. As a consequence, generations from the baseline LM are never considered to be rejected due to the \textit{\#verse} condition.

\paragraph{Perplexity.} To understand how well the model is able to leverage the information provided by a known structure, we compare the per-token perplexity of (i) PoeLM conditioned on the relevant structure descriptor, (ii) PoeLM without conditioning on any structure descriptor, (iii) the baseline LM.
We do this both in the validation set of the non-poetic corpus used for training, as well as the poem datasets used for evaluation.

Consistent with training, we insert break tokens to separate lines for both PoeLM variants. However, these special tokens are excluded from the perplexity computation to make them comparable with the baseline LM.

\subsection{Human evaluation}
\label{sec:human-evaluation}

We run a qualitative evaluation in Spanish comparing poems generated by our system and humans. Given that writing poems is also challenging for humans, we consider both poems written by actual poets as well as layman volunteers. More concretely, we take the first line of 50 poems from the DISCO dataset, and compare 3 poems generated by completing them as follows:
\begin{itemize}
    \item \textbf{Expert}: The original poem from DISCO from which the first line was extracted, authored by a renowned poet.
    \item \textbf{Layman}: Poems written by non-expert volunteers within a time limit of about 5 minutes.
    \item \textbf{PoeLM}: Poems generated by our system using the full pipeline described in \S\ref{sec:re-ranking}.
\end{itemize}

We then give these 3 poems\footnote{A 4th candidate, which we ignore when calculating the ratings, was also included for the analysis in \S\ref{sec:human-rerank}.} to human evaluators in a random order, and ask them to rank from best to worst. We report results according to two metrics: the overall rank (the percentage of times that each system has been ranked in each position), and the head-to-head comparison (the percentage of times that each system has been ranked before each other system).

All volunteers that wrote the poems, as well as those that ranked the candidates, are native Spanish speakers with university studies. While there was an overlap between both groups of volunteers, we made sure that volunteers were never asked to rank poems written by themselves. All volunteers are familiar with the fundamentals of formal verse poetry, but are not experts in the matter. Refer to Appendix \ref{app:human-evaluation} for more details.

\section{Results}

We next discuss our main results for the automatic (\S \ref{sec:eval-quant}) and human evaluation (\S\ref{sec:eval-human}).

\begin{table}[t]
\begin{small}
\begin{center}
  \addtolength{\tabcolsep}{-1.5pt}
  \begin{tabular}{rcclcc}
    \toprule
    & \multicolumn{2}{c}{Spanish} && \multicolumn{2}{c}{Basque} \\
    \cmidrule{2-3} \cmidrule{5-6}
    & PoeLM & LM && PoeLM & LM \\
    \midrule
    Correct & 30.9 & 0.0 && 23.4 & 0.0 \\
    \midrule 
    \textit{Reject due to} & & & & & \\
    \#Verse & 3.7  & 0.0 && 9.6 & 0.0 \\
    \#Slb &  17.0 & 96.6 && 34.0 & 90.3\\
    Rhyme & 13.1 & 3.4 && 11.1 & 9.7\\
    Rep. word & 31.1 & - && 19.7 & - \\
    BLEU & 4.2 & - && 2.3 & - \\
    \bottomrule
\end{tabular}%
\end{center}
\end{small}
\caption{Percentage of filtered candidates, with a breakdown for the reason of rejection. See \S\ref{sec:quant} for details.}
\label{tab:quant-stats}
\end{table}

\begin{table}[t]
\begin{small}
\begin{center}
  \addtolength{\tabcolsep}{-2.5pt}
  \begin{tabular}{rrcclcc}
    \toprule
    && \multicolumn{2}{c}{Spanish} && \multicolumn{2}{c}{Basque} \\
    \cmidrule{3-4} \cmidrule{6-7}
    && poetic & prose && poetic & prose \\
    \midrule
    \multicolumn{2}{r}{Baseline LM} & 62.7 & 15.9 && 151.1 & 24.3   \\
    \midrule
    \multirow{2}{*}{PoeLM}
    &  w/ struc & 49.5 & 11.7 && 42.5 & 10.1   \\
    & no struc & 129.5 & 18.0  && 634.2 & 81.4 \\
    \bottomrule
\end{tabular}%
\end{center}
\end{small}
\caption{Perplexity of poetic and non-poetic (prose) corpora. %
See \S\ref{sec:quant} for details.}
\label{tab:ppl-stats}
\end{table}

\subsection{Automatic evaluation}
\label{sec:eval-quant}

We report \textbf{filtering rate} results in Table \ref{tab:quant-stats}. We find that 30.9\% of Spanish poems and 23.4\% of Basque poems sampled from PoeLM meet the given constraints. While far from perfect, this means that sampling a few candidates is enough to obtain a valid poem with our approach. In contrast, none of the poems generated by the baseline LM is valid, showing that our proposed structure-aware training is critical to generate formal verse poetry with LMs. Regarding the reason for rejection, we find that the majority of candidates from PoeLM are discarded for repeating rhyming words, which the model was not directly trained to prevent.

Table \ref{tab:ppl-stats} reports the \textbf{perplexity} results.
When conditioned on structure descriptors, our model always outperforms the baseline LM, meaning that it is able to make better predictions accounting for the meter and rhyme constraints. However, when the structure descriptor is not provided, our model's perplexity is higher, presumably because the model did not see text without structure descriptors during training.%

\begin{table}[t]
\begin{small}
\begin{center}
\begin{tabular}{l*{3}{c}}
\toprule
\backslashbox{S1}{S2}
&Expert &Layman&PoeLM \\
\midrule
Expert & - & 54.0\% & 62.7\%  \\
Layman & 46.0\% & - & 60.7\%  \\
PoeLM & 37.3\% & 39.3\% & -  \\
\bottomrule
\end{tabular}
\end{center}
\end{small}
\caption{Percentage of times that system $S1$ is ranked ahead of $S2$ in the human evaluation. }
\label{tab:human_head2head}
\end{table}

\begin{table}[t]
\begin{small}
\begin{center}
\begin{tabular}{l*{3}{c}}
 \addlinespace[-\aboverulesep] 
\toprule
&1st &2nd &3rd \\ 
\midrule
Expert & 44.0\% &	28.6\% &27.3\% \\
Layman & 36.7\%&	33.3\%&	30.0\%\\
PoeLM & 19.3\%& 38.0\%&	42.7\% \\
\bottomrule
\end{tabular}
\end{center}
\end{small}
\caption{Percentage of times that each system has been ranked in each position in the human evaluation.}
\label{tab:human_ranking}
\end{table}

\subsection{Human evaluation}
\label{sec:eval-human}

We report head-to-head results in Table \ref{tab:human_head2head}, and ranking results in Table \ref{tab:human_ranking}. Human evaluators prefer poems generated by our system over those written by renowned poets in 37.3\% of the cases. Similarly, our system does better than laymen in 39.3\% of the cases.
This shows that our system is able to generate high-quality poems, which humans often prefer over poems written by other humans. This can also be seen in the ranking evaluation, as our system %
has been ranked in first position in $19.3$\% of cases, and among the first two positions in $57.3$\% of cases. 

Finally, it is surprising that layman poems are ranked above those from renowned poets nearly half of the times. We attribute this to the human evaluators themselves being laymen, leading them to prefer poems that use more plain language. This is also reflective of the subjective nature of the task, as different readers might enjoy poetry differently. %

\section{Analysis}

We further analyze our system by quantifying at which portion of the poem its perplexity gain is highest (\S\ref{sec:perpgain}), experimenting with manual re-ranking  (\S\ref{sec:human-rerank}), and looking at some sample poems (\S\ref{sec:samples}).

\begin{figure}[t]
\includegraphics[width=\linewidth]{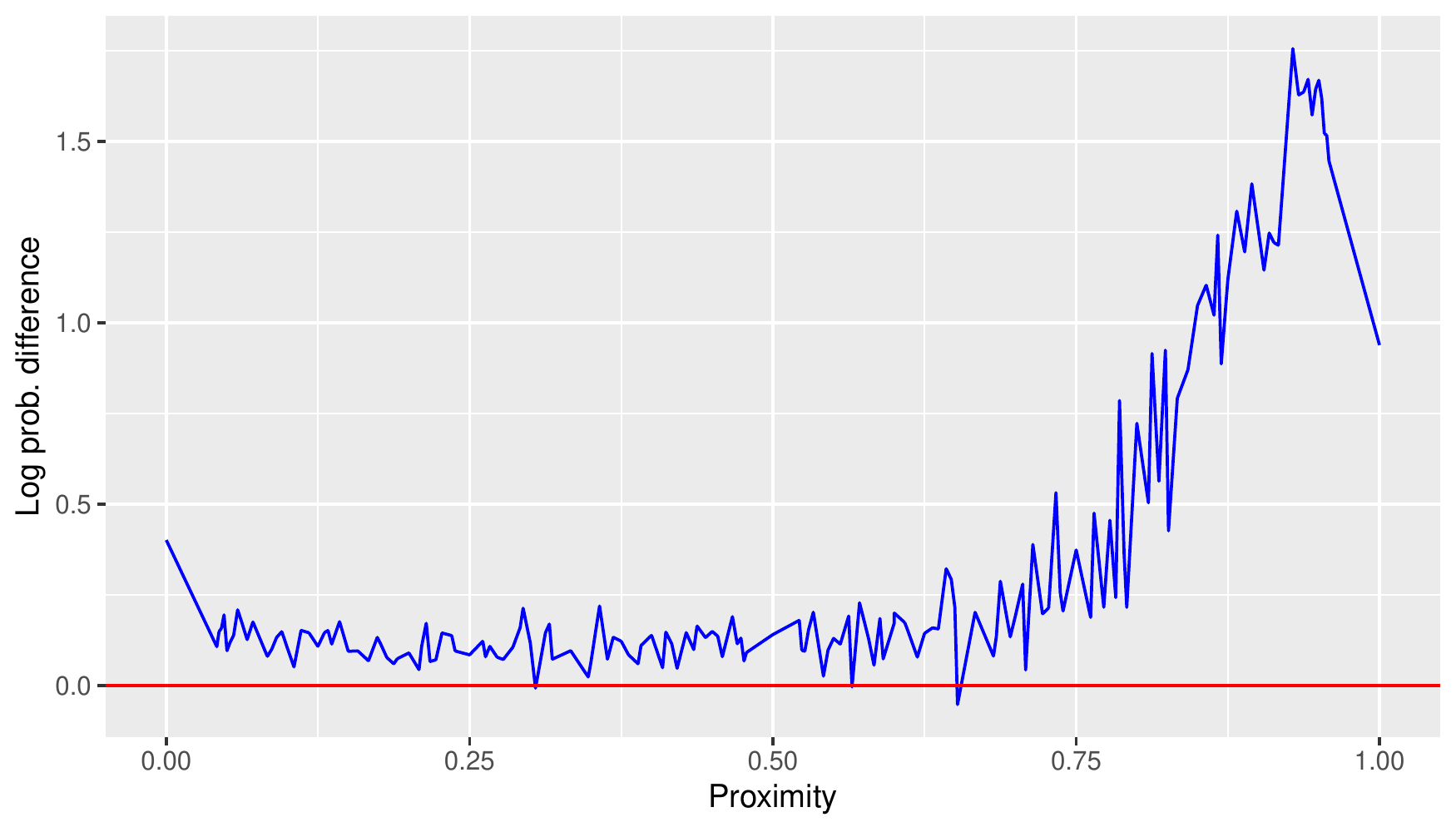}
\caption{Interpolated advantage in log probability of our model compared to a regular LM over the Spanish mC4 validation set, as a function of normalized proximity to the next specified rhyme token. See \S\ref{sec:perpgain} for details. %
}
\label{fig:logprobadvantage}
\end{figure}

\subsection{Perplexity gain}
\label{sec:perpgain}

We quantify the predictive advantage of our system as a function of the distance to the next rhyme word. To this end, we plot the difference in token-wise log probabilities between our model and the baseline LM as a function of proximity to the next rhyme word, interpolated between 0 and 1. We only consider lines with 15 to 25 tokens.

As shown in Figure \ref{fig:logprobadvantage}, our model's advantage is greatest near the rhyme word. This is not surprising, as there is less uncertainty towards the end of the line when the meter and rhyme are known. We observe a downward spike towards the end, that may initially seem counter-intuitive. We hypothesize that, since the rhyme word will often be split into multiple tokens, by the time the first tokens of the rhyme word are known the regular LM will be quite sure of what the word is, meaning that the advantage of knowing the rhyme is lower.

\begin{table}[t]
\begin{small}
\begin{center}
\begin{tabular}{l*{4}{c}}
\toprule
\multirow{2}{*}{\backslashbox{S1}{S2}}
&Exp. &Lay. &PoeLM & PoeLM\\
&&&&\textit{\small +rerank} \\ 
\midrule
PoeLM & 37.3& 39.3 & - & 26.0 \\
PoeLM & \multirow{2}{*}{41.3}&  \multirow{2}{*}{42.7} &  \multirow{2}{*}{38.0}&  \multirow{2}{*} - \\
\textit{+rerank} &&&&\\
\bottomrule
\end{tabular}
\end{center}
\end{small}
\caption{Percentage of times that system $S1$ is ranked ahead of $S2$ in the human evaluation. Since the candidate chosen by a human annotator among the top 6 candidates will sometimes be the same as the top candidate, there can be ties, and thus the head-to-head percentages do not add up to 100.  
}
\label{tab:re-rank}
\end{table}

\subsection{Manual re-ranking}
\label{sec:human-rerank}

A potential application of automatic poetry generation is helping (rather than replacing) humans when writing poems. As a first approximation, we ask our volunteers to manually choose a poem among the top 6 candidates generated by our system.\footnote{We take the top three candidates with and without BLEU filtering to generate this list.} The resulting poem was considered as part of the human evaluation described in \S\ref{sec:human-evaluation}, and compared to the other 3 systems.

Table \ref{tab:re-rank} reports the head-to-head performance of our model with and without manual re-ranking. As expected, the re-ranked model performs better, beating the poems generated by laymen in 42.7\% of cases, as opposed to 39.3\% for the base system. However, the base system beats manual re-ranking in 26\% of cases, meaning that human evaluators often prefer the top candidate automatically selected by the system over the one manually selected by another human. This means that there is a considerable disagreement across annotators, which is reflective of the subjective nature of the task.

\begin{table*}[t]
\centering
\begin{small}
\begin{tabular}{ccc}
\toprule 
Generated Poem && English translation \\
\cmidrule{1-1} \cmidrule{3-3} 
\makecell{
Siento otro Yo que contra mí se empeña,\\
un Yo para el que no debo luchar,\\
un Yo al que no debo acompañar,\\
un Yo que a menudo me condena.} &&
\makecell{I feel another Self that is set against me,\\
a Self for which I must not fight,\\
a Self that I should not accompany,\\
a Self that often condemns me.
}
\\ \midrule
\makecell{Casta su faz, bajo la blanca toca,\\
su faz es dulce,es bella sin cesar,\\
su faz es hermosa como un jaguar,\\
su faz es divina como la roca.} &&
\makecell{Chaste is her face, under her white cap,\\
her face is sweet,  relentlessly gorgeous,\\
her face is beautiful like a jaguar,\\
her face is divine like rock.}
\\ \midrule
\makecell{Nada más. De aquel sueño, que pasó como un ala,\\
arranco todo lo que había en mi mente,\\
todo lo que me atormentaba y no cala,\\
lo que callo en el interior de mi presente.} &&
\makecell{Nothing more. Of that dream, that passed like a wing,\\
it tore everything from my mind,\\
all that tormented me and doesn't seep through, \\
which I keep quiet inside my present. }\\
\midrule 
\makecell{Es inútil que luche por quitarme del pecho,\\
que niegue repetidamente mis opiniones,\\
que trague de nuevo mi entusiasmo deshecho,\\
que rechace de nuevo todas mis negaciones.} &&
\makecell{Fighting to get it off my chest is futile,\\
that I repeatedly deny my emotions,\\
that I once again swallow my undone enthusiasm,\\
that I once again reject all my negations.
}\\
\midrule 
\makecell{Del Monserrat en la penumbra undosa,\\
Del Monserrat silente en el Solar,\\
Del Monserrat dolido en el remar,\\
Del Monserrat cautivo en la prosa.} &&
\makecell{Of the Monserrat in the gloomy twilight,\\
Of the Monserrat, silent in  sunlight,\\
Of the Monserrat, pained in paddling,\\
Of the Monserrat, captive in prose.}\\
\bottomrule 
\end{tabular}
\end{small}
\caption{Spanish poems generated by our method, given five lines selected at random from the dataset. The five poems have been manually selected from the top six candidates generated by the system for each line, with no other form of \textit{cherry-picking}. See Appendix \ref{sec:all-poems} for the full list of six candidate poems.}
\label{tab:examples-spanish}
\end{table*}

\begin{table*}[t]
\centering 
\begin{small}
\begin{tabular}{ccc}
\toprule 
Generated Poem && English translation \\
\cmidrule{1-1} \cmidrule{3-3}
\makecell{
Gu biok lagun handiak gara,\\
anaia,aita,semea,\\
 eta bion ideologia,\\
gure identitatea,\\
konpartitzen dugu.Batzuetan,\\
zaila da bat esatea,\\
besteak ulertzea,benetan,\\
zein ahula den bestea.
}
&& \makecell{
The both of us are great friends,\\
brother,father,son,\\
and our ideology,\\
our identity,\\
is shared. Sometimes,\\
it is hard to say one,\\
to understand others, truly,\\
how weak others are.
}
\\ \midrule
\makecell{Nahiz kulunpio pila bat egon,\\
eguzkiak sikiera,\\
aukera du ondo goxatzeko,\\
eta ez beti gainera,\\
baita asteko egun denetan,\\
baita hemendik aurrera,\\
ilargi erdiko orduetan,\\
eta hori da ederra.
}&& \makecell{Even though there are many swings,\\
at least the sun,\\
has a chance to enjoy,\\
not always,\\
also during every day of the week,\\
and, from now on,\\
during the moon hours,\\
and that is beautiful.}\\

\bottomrule 
\end{tabular}
\end{small}
\caption{Basque  poems generated by our method, given two lines selected at random from the dataset. The poems have been manually selected from the top six candidates generated by the system for each line, with no other from of \emph{cherry-picking}.  See Appendix \ref{sec:all-poems} for the full list of six candidate poems.}
\label{tab:examples-basque}
\end{table*}

\subsection{Sample poems}
\label{sec:samples}
Tables \ref{tab:examples-spanish} and \ref{tab:examples-basque} show some example poems generated by our system in Spanish and Basque. 
The poems were generated by providing the first line along with the structure descriptor to the system, and manually selecting a candidate among the top six. Five lines were selected at random from the evaluation in Spanish, and two for Basque. The full list with the six candidates is given in Appendix \ref{sec:all-poems}. No \emph{cherry-picking} was done, except to choose one poem among the six candidates per line.

We observe that the system is capable of generating coherent poems covering varied topics. For example, regarding the Spanish poems, the first, third and fourth Spanish poems cover themes of inner conflict, the second one describes a person's beauty, and the last is about an abbey called Monserrat. Note that the theme is implicit in the first line, and mirrors the typical topics of Spanish sonnets of the time. Regarding the Basque poems, the themes are friendship and swings in a park, also mirroring the themes used in contemporary spontaneous poetry contests in the dataset.

\label{sec:sample-poems}

\section{Related work}

\label{sec:poetrygen}

We next review relevant literature in poetry generation (\S\ref{sec:poetrylit}), as well as controllable generation (\S\ref{sec:controllablelit}). 

\subsection{Poetry generation}
\label{sec:poetrylit}

\textbf{Retrieval based approaches.} Early work in poetry generation focused on rule-base methods, which generate text according to predefined rules that ensure the desired structure is followed \citep{wasp-gervas02,tralalyrics-oliveira07}. A popular approach is to fill templates with text extracted from existing poems \citep{fullface-colton12,poetryme-oliveira12,mpoetryme-oliveira17}. This makes it easy to control poetic structure, since the meter and rhyme schemes of the text pieces can be annotated in advance and combined accordingly when filling the templates. However, the diversity and creativity of these approaches is limited.

\textbf{Neural poetry generation.} More recently, there has been work on applying neural text generation to poetry. A popular approach is to train a finite-state acceptor (FSA) that ensures all accepted sequences obey the required structure, which is then used to guide a recurrent neural network (RNN) through rejection sampling \citep{ghazvininejad-etal-2016-generating, ghazvininejad-etal-2018-neural, hopkins-kiela-2017-automatically}. However, these methods require some form of lyrical or poetic text to train the RNN or the FSA, and they must generate text right-to-left in order to respect rhyme sounds, as the model has no concept of planning. Additionally, a new FSA has to be trained for each desired poem structure. \citet{lau-etal-2018-deep} augment an RNN with a pentameter model and learn the meter and rhyme constraints of sonnets in a supervised way from a sonnet corpus. They then generate poem lines right-to-left, to alleviate the model's lack of planning. \citet{van-de-cruys-2020-automatic} trains an encoder-decoder RNN on prosaic text to generate each line right-to-left conditioned on the previous one, and applies constraints when decoding to ensure the generated text adheres to a rhyme scheme and consistent topic. However, their system cannot enforce a specific syllabic meter.

Multiple works focus on neural poetry generation for the Chinese language, applying techniques such as reinforcement learning \citep{yi-etal-2018-automatic} or planning \citep{wang-etal-2016-chinese}.
In Chinese, one character corresponds to a syllable, but meter is governed by tonal constraints. Most of the reviewed works assume that, with a sufficiently large corpus, the model should be able to learn the implicit tonal structure of poetry \cite{wang-etal-2016-chinese,zhang-etal-2017-flexible,liu2018multimodalpoetry}. \citet{yeh2019rhyming} concatenate tonal information to the character embeddings of an LSTM to create a model that is more phonologically compliant.

 Notably, current neural methods capable of controlling both syllable count and rhyme scheme require some form of poetic corpus to train, and usually generate text right-to-left to alleviate a lack of planning when generating rhymes.

\subsection{Controllable generation}
\label{sec:controllablelit}

Similar to our approach, several works attempt to control the generated output by augmenting the training data with tags. \citet{keskarCTRL2019} augment the training corpus of a LM with codes automatically extracted from metadata. Some works in machine translation explore augmenting the training data in order to control the politeness \citep{sennrich-etal-2016-controlling}, domain \citep{kobus2017}, or length \citep{lakew-etal-2019-controlling} of generated translations. \citet{schioppa-etal-2021-controlling} experiment with vector-valued additive tags in order to control multiple attributes of the generated text at once. However, all of these systems use tags that only broadly specify the length, domain or style of the text to generate. In contrast, our model is conditioned on a very specific meter and rhyme scheme that the text must follow.

\section{Conclusions and future work}

In this work, we present an unsupervised approach to generate formal verse poetry. We identify and extract the latent structure in non-poetic corpora, and feed this information along with the text to a transformer LM, allowing us to control the structure of the text at generation time. Our system is capable of generating formal verse poetry with flexible meter and rhyme schemes, without requiring any sort of poetic text to train. The required structure can be easily altered by changing the descriptor, allowing us to generate different types of poetry without needing to re-train the system. 
Automatic and human evaluations show that our model learns to leverage the provided structure information to better predict the text, and is capable of generating short poems that are often preferred to those created by a human. 

In future work, we would like to extend our framework to be able to control other aspects of the generated text in addition to meter and rhyme. 

\section*{Limitations}

Given that our method requires tagging the implicit meter and rhyme of the training corpus, we are limited by the quality of available syllabization and rhyme detection systems. While rule-based systems with a low error rate are easy to create for languages such as Spanish or Basque, this is not the case for English, which is why we did not train an English version of our system. However, our approach is independent of the used syllabization and rhyme detection process, and could be readily applied on top of any system with a low error-rate.

Additionally, our Spanish syllabization system has no concept of synalephas, where two syllables can be merged into one when one word ends in a vowel and the next starts with one. This means that our system will never use this Spanish literary device when generating poems.

\section*{Acknowledgements}
Aitor Ormazabal, Aitor Soroa and Eneko Agirre were partially supported by the Basque Government (IXA excellence research group IT1343-19). Aitor was supported by a doctoral grant from the Spanish MECD.

We would like to thank Ainara Estarrona, Begoña Altuna and Itziar Gonzalez-Dios for their assistance in defining literary terminology, and Edward Yao for his help translating poems.

\bibliography{anthology,custom}
\bibliographystyle{acl_natbib}

\appendix

\section{Structure descriptors}
\label{sec:structure-descriptors}

The process of extracting the meter descriptors from a regular corpus and creating the augmented corpus consists of four steps:

\begin{enumerate}
    \item First, we split the text into phrases according to the following set  delimiters: \_-?"!,:’‘()[].\{\}`;»«><'. We do this so that the phrases, which will correspond to lines in our generated poems, end at natural stopping points in speech. Additionally, we randomly merge each phrase with the next or the next two phrases, with probabilities $0.15$ and $0.05$, respectively, so that the model can generate verses that contain these special characters.
    \item Second, we syllabize each phrase and extract the rhyme class of its final word, using our FOMA transducers. 
    \item Third, we split the text into blocks of $n$ phrases, where $n$ is sampled uniformly between $3$ and $10$. For each block, we construct a meter descriptor from the syllable count and rhyme class of each phrase. The descriptor begins with a \texttt{<PREF>} token and ends with a \texttt{</PREF>} token, and a pair of tokens of the form \texttt{<LEN\_X> <CLS\_Y>} for each phrase, where $X$ is the syllable count and $Y$ is the rhyme class. In $15\%$ of cases, the rhyme class is replaced with a special \texttt{<CLS\_UNK>} class, which allows us to leave the rhyme of certain verses unspecified when generating. Additionally, when there is a paragraph boundary (line break) in the text, we insert a \texttt{<SEP>} token in the corresponding position. 
    \item Fourth, we construct the corpus interleaving the meter descriptors in between the corresponding blocks of text. Additionally, we insert a \texttt{<BRK>} token in between phrases in the actual text. The \texttt{<BRK>} token lets us  clearly see where the model has intended to end a phrase, allowing us to split the generated poems into verses easily.
\end{enumerate}

\section{Human evaluation}
\label{app:human-evaluation}

As described in \S\ref{sec:human-evaluation}, we extract the first line from fifty real poems in the DISCO dataset, and compare the poems obtained by completing them. Specifically, for each case, we present the four candidate completions in a randomized order, and ask a human volunteer evaluator to rank them from best to worst. Additionally, the rhyme scheme that the poem should follow is provided, so that the evaluator can take it into account when judging a poem (our systems' poems and the gold standard never contain rhyme mistakes, but the poems generated by human annotators might). Each ranking is done by three different annotators. 

All volunteers that participated in writing and ranking poems are familiar with the fundamentals of formal verse poetry, but are not experts in the matter. To obtain the "Layman" poems, volunteers were asked to complete a poem within a time limit of 5 minutes. However, this time limit was not enforced, and it was slightly exceeded in some cases.

\section{Full list of poems}
\label{sec:all-poems}

Tables \ref{tab:fullpoemsspanish} and \ref{tab:fullpoemsbasque} show the full list of the six candidate poems generated by our system for each of the five lines in Spanish and two lines in Basque chosen at random from their respective datasets. No cherry-picking was done on these tables. From these poems, we did select one for each line manually, as shown in Tables \ref{tab:examples-basque} and \ref{tab:examples-spanish} in the main body of the paper (see \S\ref{sec:samples}).

\begin{table*}[t]
\centering
\begin{small}
\begin{tabular}{cc }
\toprule 
\makecell{Siento otro Yo que contra mí se empeña,\\
Siento otro Yo que no quiere cambiar,\\
Siento otro Yo que no sabe amar,\\
Siento otro Yo que no se condena. }\\  
 \midrule
\makecell{ Siento otro Yo que contra mí se empeña,\\
siento otro Yo que no quiero amar,\\
siento otro Yo que no puedo hablar,\\
siento otro Yo que no me enseña. }\\  
 \midrule
\makecell{ Siento otro Yo que contra mí se empeña,\\
y otro Yo que me quiere dominar,\\
y otro Yo que no me quiere amar,\\
que me condena y que me condena. }\\  
 \midrule
\makecell{ Siento otro Yo que contra mí se empeña,\\
un Yo para el que no debo luchar,\\
un Yo al que no debo acompañar,\\
un Yo que a menudo me condena. }\\  
 \midrule
\makecell{ Siento otro Yo que contra mí se empeña,\\
o una voz que me hace suspirar,\\
o una risa que me hace llorar,\\
o un grito que me quita la pena. }\\  
 \midrule
\makecell{ Siento otro Yo que contra mí se empeña,\\
un Yo que no me deja amamantar,\\
un Yo que me ensena a no llorar,\\
un Yo que me enseña lo que pena. }\\  
 \midrule
 \midrule
\makecell{ Casta su faz, bajo la blanca toca,\\
vestida de dos trpitos de azahar,\\
Casta su faz,bajo la blanca tocar,\\
vestida de dos trpitos de su boca. }\\  
 \midrule
\makecell{ Casta su faz, bajo la blanca toca,\\
carvaja y danseá del palomar,\\
Casta su faz,bajo la blanca tocar,\\
carvaja y danseá de la roca. }\\  
 \midrule
\makecell{ Casta su faz, bajo la blanca toca,\\
casta su faz sin tabiques que tapar,\\
casta su faz cuando quiere cabalgar,\\
casta su faz con guarnición de boca. }\\  
 \midrule
\makecell{ Casta su faz, bajo la blanca toca,\\
su faz es dulce,es bella sin cesar,\\
su faz es hermosa como un jaguar,\\
su faz es divina como la roca. }\\  
 \midrule
\makecell{ Casta su faz, bajo la blanca toca,\\
ennoblece este cantar popular,\\
casta su faz,bajo la blanca tocar,\\
ennoblece este cantar de boca. }\\  
 \midrule
\makecell{ Casta su faz, bajo la blanca toca,\\
no dejaba errar;casta su mirar,\\
no dejaba turbar;casta su hablar,\\
no dejaba hacer;casta su boca. }\\  

\bottomrule 
\end{tabular}
\end{small}
\end{table*}

\begin{table*}[t]
\centering
\begin{small}
\begin{tabular}{cc }
\toprule 
\makecell{ Nada más. De aquel sueño, que pasó como un ala,\\
voló el agua como un ave frugalmente,\\
De aquel sueño,que pasó como una bala,\\
voló el viento como un átomo ardiente. }\\  
 \midrule
\makecell{ Nada más. De aquel sueño, que pasó como un ala,\\
no me devolviera la pálida piel caliente,\\
De aquel sueño,que pasó como una bala,\\
no me devolviera la herida en la frente. }\\  
 \midrule
\makecell{ Nada más. De aquel sueño, que pasó como un ala,\\
arranco todo lo que había en mi mente,\\
todo lo que me atormentaba y no cala,\\
lo que callo en el interior de mi presente. }\\  \midrule
\makecell{ Nada más. De aquel sueño, que pasó como un ala,\\
salió un pensamiento que viniera al frente,\\
un pensamiento que pasó como una bala,\\
salió un rayo que se encendió en mi mente. }\\  
 \midrule
\makecell{ Nada más. De aquel sueño, que pasó como un ala,\\
como un beso que se derrumbó de mi mente,\\
y que se fue al pasado como una bala,\\
y desapareció del mapa como un ente. }\\  
 \midrule
\makecell{ Nada más. De aquel sueño, que pasó como un ala,\\
un ave se paró sola en un parapente,\\
De aquel sueño,que pasó como una bala,\\
un viento se detuvo en el mar de enfrente. }\\  
 \midrule
 \midrule
\makecell{ Es inútil que luche por quitarme del pecho,\\
Es muy difícil que yo huya de mis traiciones,\\
Es demasiado tarde para lo que he hecho,\\
Es demasiado tarde para mis decisiones. }\\  
 \midrule
\makecell{ Es inútil que luche por quitarme del pecho,\\
que niegue repetidamente mis opiniones,\\
que trague de nuevo mi entusiasmo deshecho,\\
que rechace de nuevo todas mis negaciones. }\\  
 \midrule
\makecell{ Es inútil que luche por quitarme del pecho,\\
que no sea la tiranía de las pasiones,\\
que se ría de mí de lo que yo le he hecho,\\
Es inútil que se enoje por mis canciones. }\\  
 \midrule
\makecell{ Es inútil que luche por quitarme del pecho,\\
es inútil que llore por tus provocaciones,\\
es inútil que te diga qué es lo más hecho,\\
es inútil que afirme mis acusaciones. }\\  
 \midrule
\makecell{ Es inútil que luche por quitarme del pecho,\\
que luche por alcanzarme con sus oraciones,\\
que luche por bajarme del caballo derecho,\\
que me meta en mi cama con sus peticiones. }\\  
 \midrule
\makecell{ Es inútil que luche por quitarme del pecho,\\
que me refugie en mi casa de ilusiones,\\
que me grite a voces que quiero y no hecho,\\
que me arregle los días sin palpitaciones. }\\  

\bottomrule 
\end{tabular}
\end{small}
\end{table*}

\begin{table*}[t]
\centering
\begin{small}
\begin{tabular}{cc }
\toprule 

\makecell{ Del Monserrat en la penumbra undosa,\\
Del Monserrat en la luz crepuscular,\\
Del Monserrat en la vida de Aznar,\\
Del Monserrat en la noche ansiosa. }\\  
 \midrule
\makecell{ Del Monserrat en la penumbra undosa,\\
Del Monserrat en la niebla uncular,\\
Del Monserrat en la luna anular,\\
Del Monserrat en la noche brumosa. }\\  
 \midrule
\makecell{ Del Monserrat en la penumbra undosa,\\
Del Monserrat en la noche un lugar,\\
Del Monserrat en la luz un despertar,\\
Del Monserrat en el sol una rosa. }\\  
 \midrule
\makecell{ Del Monserrat en la penumbra undosa,\\
Del Monserrat con los dedos sin borrar,\\
Del Monserrat con las ganas de cantar,\\
Del Monserrat de la roca filosa. }\\  
 \midrule
\makecell{ Del Monserrat en la penumbra undosa,\\
Del Monserrat silente en el Solar,\\
Del Monserrat dolido en el remar,\\
Del Monserrat cautivo en la prosa. }\\  
 \midrule
\makecell{ Del Monserrat en la penumbra undosa,\\
Del Monserrat cuando hay que claudicar,\\
Del Monserrat hasta que hay que llorar,\\
Del Monserrat con mantilla larosa. \\
}\\  
 \bottomrule

\end{tabular}
\end{small}
\caption{Full list of six candidate poems for Spanish generated by our system for the five lines chosen at random in the Spanish dataset. No cherry-picking was done on this table. From these poems, we did select one for each line manually, as shown in Table \ref{tab:examples-spanish} in the main body of the paper.
}
\label{tab:fullpoemsspanish}
\end{table*}

\begin{table*}[t]
\centering
\begin{small}
\begin{tabular}{cc }
\toprule 
\makecell{Gu biok lagun handiak gara,\\
anaia,aita,semea,\\
 eta bion ideologia,\\
gure identitatea,\\
konpartitzen dugu.Batzuetan,\\
zaila da bat esatea,\\
besteak ulertzea,benetan,\\
zein ahula den bestea.}\\  
 \midrule
\makecell{Gu biok lagun handiak gara,\\
bata zinemazalea,\\
bestea musikaria bera,\\
biek dugu lanbidea,\\
Denok egiten dugu aurrera,\\
dena da inportantea,\\
dena dago egiteko era,\\
izaten da idazlea.}\\  
 \midrule
\makecell{Gu biok lagun handiak gara,\\
argazkilari zalea,\\
eta asko gustatzen zait bera,\\
nire bikotekidea,\\
argazki kamera eskuetan,\\
ezaguna izatea,\\
gogoratzen dudan argazkia,\\
lagunekin ikustea.}\\  
 \midrule
\makecell{Gu biok lagun handiak gara,\\
Esne Beltzako kidea,\\
anai zaharra eta arreba,\\
Jon eta nire semea,\\
eta bi panpin gaude honetan,\\
Oihan Vega da bestea,\\
Hesian taldeko kantaria,\\
Sorotan Bele taldea.}\\  
 \midrule
\makecell{Gu biok lagun handiak gara,\\
katuak ardi taldea,\\
tomateak eta haragia,\\
soja eta laborea,\\
oilaskoak fruitua lortzeko,\\
lekalearen jeztea,\\
baratzea animalietan,\\
pinpilinpauxa zaintzea.}\\  
 \midrule
\makecell{Gu biok lagun handiak gara,\\
baina gure bikotea,\\
itxura guztien arabera,\\
ez da bideko jendea,\\
Bidaia ez da gauza berria,\\
ez da hala esatea,\\
Baina batzuetan,batzuetan,\\
izan daiteke jabea.}\\  
 
 \bottomrule

\end{tabular}
\end{small}
\end{table*}

\begin{table*}[t]
\centering
\begin{small}
\begin{tabular}{cc }
\toprule 
\makecell{Nahiz kulunpio pila bat egon,\\
galtza belauna behera,\\
gero eskailerak,hortz parea,\\
sokak,besoak atzera,\\
kautxuzko motxilak bizkotxetan,\\
 eta astindu ederra,\\
bularra,pultsuko aurpegia,\\
izterra eta izterra.}\\  
 \midrule
\makecell{Nahiz kulunpio pila bat egon,\\
ez da oso gauza bera,\\
autoa hartu eta kotxea,\\
leku batetik bestera,\\
beste leku batera sartzeko,\\
jendearekin batera,\\
egiten dugun bide horretan,\\
kotxeak izorratzera.}\\  
 \midrule
\makecell{Nahiz kulunpio pila bat egon,\\
aspiranteen antzera,\\
gorputza sentsuala da ia,\\
betirako izaera,\\
eta erretzen duzu airea,\\
ez zara esterilera,\\
kaka egiten duzu oinetan,\\
likidoa da gainera.}\\  
 \midrule
\makecell{Nahiz kulunpio pila bat egon,\\
eguzkiak sikiera,\\
aukera du ondo goxatzeko,\\
eta ez beti gainera,\\
baita asteko egun denetan,\\
baita hemendik aurrera,\\
ilargi erdiko orduetan,\\
eta hori da ederra.}\\  
 \midrule
\makecell{Nahiz kulunpio pila bat egon,\\
komunera lasaitzera,\\
komunera salto egiteko,\\
irakurtzera,jotzera,\\
txorimaloetaz gozatzea,\\
konpainia egitera,\\
burrunba burrunbatsuaz janztea,\\
koadrilan afaltzera.}\\  
 \midrule
\makecell{Nahiz kulunpio pila bat egon,\\
bada zerbaiten plazera,\\
dena zure eskura uztea,\\
dakizuna arabera,\\
izan zaitez supergizakia,\\
izan nahi izatera,\\
hau da,zuhaur,dena emateko,\\
munduarekin batera.\\
}\\  

 \bottomrule

\end{tabular}
\end{small}
\caption{Full list of six candidate poems for Basque generated by our system for the two lines chosen at random in the Basque dataset. No cherry-picking was done on this table. From these poems, we did select one for each line manually, as shown in Table \ref{tab:examples-basque} in the main body of the paper.
}
\label{tab:fullpoemsbasque}
\end{table*}

\end{document}